# Analyzing Breast Cancer Survival Disparities by Race and Demographic Location: A Survival Analysis Approach

Ramisa Farha  Joshua O. Olukoya
Computer Science Department
Morgan State University, MD 21251
Rafar2@morgan.edu, Joolu5@morgan.edu

## Abstract

This study employs a robust analytical framework to uncover patterns in survival outcomes among breast cancer patients from diverse racial and geographical backgrounds. This research uses the SEER 2021 dataset to analyze breast cancer survival outcomes to identify and comprehend dissimilarities. Our approach integrates exploratory data analysis (EDA), through this we identify key variables that influence survival rates and employ survival analysis techniques, including the Kaplan-Meier estimator and log-rank test and the advanced modeling Cox Proportional Hazards model to determine how survival rates vary across racial groups and countries. Model validation and interpretation are undertaken to ensure the reliability of our findings, which are documented comprehensively to inform policymakers and healthcare professionals. The outcome of this paper is a detailed version of statistical analysis that not just highlights disparities in breast cancer treatment and care but also serves as a foundational tool for developing targeted interventions to address the inequalities effectively. Through this research, our aim is to contribute to the global efforts to improve breast cancer outcomes and reduce treatment disparities.

**Keywords**: Breast Cancer, Survival Analysis, Racial Disparities, Demographics, Cox Proportional Hazards model, Kaplan-Meier estimator, Log-rank test, Statistical analysis.

## I. Introduction

Breast cancer has been a global health concern and one of the leading causes of death among women. Despite substantial advances in therapy and early detection, survival rates are unevenly distributed with notable variations across racial and geographic boundaries for a long time. Disparities are caused by various complex interactions of factors such as socioeconomic class, access to quality healthcare, sometimes genetic predispositions, and awareness of early detection. These disparities highlight the most critical need for targeted interventions to enhance outcomes across all groups of people. The ultimate purpose is to examine breast cancer survival outcomes through a thorough analytical methodology. Breast cancer patients' survival rates are impacted by a very complex interaction of factors such as socioeconomic standing, hereditary characteristics, the accessibility and availability of healthcare services, and the timing of diagnosis and treatment. For example, delayed diagnosis—often due to a lack of knowledge or insufficient access to early screening—significantly lowers survival chances. Similarly, systemic inequalities in healthcare systems disproportionately harm marginalized areas, worsening survival gaps. These facts highlight the need for investigation of how race and geographic location influence survival outcomes for the development of targeted, data-driven interventions addressing these disparities. This study investigates survival rates across several ethnic and geographic groups using the SEER (Surveillance, Epidemiology and End Results) 2021 dataset. The study uses exploratory data analysis (EDA) and advanced statistical approaches for example Kaplan-Meier estimator, log-rank test and Cox Proportional Hazards model to identify and quantify the most essential variables influencing survival outcomes. Beyond statistical modeling, this study investigates the impactfulness of breast cancer awareness on early detection stages and survival rates. Integrating thorough data analysis and also model validation yields practical insights to the underlying causes of survival differences. These findings actually intended to help policymakers and also healthcare practitioners devise targeted initiatives that address systemic disparities, increasing early identification, and improvement in treatment outcomes for marginalized populations. Following that, this study also intends to contribute to global efforts for reducing inequities in breast cancer care and improve survival outcomes for women worldwide. The study's findings will provide not only the basis for future research but also the development of effective solutions to close gaps in breast cancer treatment and care.

## II. Related works

The study examines the most common reasons of ethnic and racial differences in breast cancer mortality and presents the most prevalent strategies on how to lower them. Inequalities in breast cancer mortality are too complicated and diverse, resulting from a mix of patient, provider, and health-care system characteristics. Patient variables include a lack of insurance, and sometimes testing anxiety, hesitation in seeking treatment, and not favorable tumor features. Mostly Inadequate assessment, poor follow-up of erroneous screening tests, and failure to adhere to guideline-based therapies are all provider-related problems. High coinsurance requirements, a lack of a mainstay of care, treatment fragmentation, and unequal allocation of treatment and screening resources are all factors in the health-care system. These factors work together and with larger social and economic determinants to cause and maintain inequities in breast cancer mortality [2].

The research addressed ethnic and racial disparities in breast cancer diagnosis and treatment. The study showed that American Indian, Southeast Asian, Black, South Asian, Pacific Islander, and Hispanic women were less than White women to be diagnosed with early-stage breast cancer. In addition, among those who were diagnosed early on, Hispanic, American Indian, Pacific Islander, and Black women had higher rates of mortality than White women. Within the late stages, Black women had greater fatality rates than White women. These discrepancies persisted across distinct genetic subtypes and disease stages, emphasizing the critical need for additional research and intervention to address and eliminate inequities in breast cancer early identification and survival [3].

This study explored the correlation between ethnicity or race, individuals socioeconomic level, and breast cancer stage upon diagnosis. The study discovered that Black and Hispanic women were more inclined than White women to be diagnosed as having early-stage breast cancer, even after accounting for distinct socioeconomic factors like income and education. The research also discovered that the approach of disease detection was a significant factor in racial/ethnic inequalities in stage at diagnosis. Women who discovered cancer spontaneously or through a clinical breast examination had a lower chance of receiving an early-stage diagnosis compared to those who were detected through mammograms for screening. The findings indicate that, while individual socioeconomic variables may play a role in breast cancer disparities, other factors such as access to screening and cultural factors may also be significant. Additional research is required to fully understand the numerous factors that contribute to racial and ethnic discrepancies in breast cancer stage at diagnosis and survival [4].

The authors of this study work investigate the recurrent issue of racial disparities in breast cancer outcomes, which is a complex phenomenon with social and structural causes. Despite general improvements in breast cancer mortality rates, African American women continue to die at higher rates than white women. This gap underlines the critical need for additional research into the social determinants of health that contribute to these unequal outcomes. It investigates numerous factors that could contribute to these differences, including socioeconomic position, access to adequate healthcare, and institutional racism. Income, education, and occupation are all socioeconomic characteristics that might impact a woman's access to preventative care, early identification, and timely treatment. Furthermore, systematic racism can erect barriers to care, resulting in inequities in diagnosis, treatment, and survival. The authors underlined the need of tackling social determinants of health for diminishing racial disparities in the outcomes of breast cancer. This includes expanding access to healthcare quality, raising health literacy and addressing structural racism. We can achieve health equity for all women by using a holistic approach that takes into account both biological and social factors [5].

In this study, the authors undertook a thorough assessment into the ongoing racial differences in breast cancer survival between Black and White women. Despite substantial advances in early identification and treatment options, Black women continue to have disproportionately higher mortality rates than their White counterparts. The study, a systematic review and meta-analysis, discovered that these inequalities exist across all major tumor subtypes, along with hormone receptor-positive and HER2-negative breast cancer showing the greatest significant difference. These findings highlight the critical need for a diversified strategy to address the complex interplay of social, economic, and institutional issues that lead to unequal outcomes. By implementing targeted treatments that address these underlying determinants of health, we may strive for health equity and improve breast cancer outcomes for Black women [6].

This paper looked at the racial disparities in breast cancer mortality among non-Hispanic White and non-Hispanic Black in Atlanta and Georgia and found Black women have much higher incidence of breast cancer-related death than White women. This differential is due to a complex interplay of causes, including the diagnosis of more aggressive tumor forms, less appropriate therapy, and socioeconomic differences. Notably, Black women with high socioeconomic position had the highest risk of mortality, implying that variables other than socioeconomic status contribute to these differences. These findings highlight the need for more study and identify the underlying reasons of racial disparities and create tailored interventions to address special needs of Black women having breast cancer, with the goal of improving their health outcomes and closing racial gap in breast cancer mortality [7].

This study looked at racial discrepancies in breast cancer survival rates of women in Brazil's public healthcare system. The results show large inequalities among women of different racial or skin color groups. Black women had a lower 5-year overall survival rate 64% against 74% for white women. Furthermore, Black women had 24% higher risk of mortality than White women. Such differences persisted after accounting for age, stage of disease, therapy and socioeconomic position emphasizing the critical need to address inequities and enhance healthcare equity for Black women in Brazil. Potential interventions include focused outreach and also education programs to raise knowledge of breast cancer risk factors and the need for early identification and treatment, as well as efforts to remove obstacles for excellent healthcare access [8].

This study paper investigates for alarming racial differences in breast cancer mortality in Black women in the United States. Despite having lower prevalence of breast cancer, Black women have a higher chance of death than White women. Such a difference became apparent mostly in the 1980s, when mammography screening and adjuvant endocrine therapy were widely adopted. It concludes that discrepancies in access to these therapies, combining a higher prevalence of aggressive breast cancer subtypes among Black women, contribute to inferior outcomes. Black women are likely to have poor coverage, which limits access to life-saving therapies. Thus, result in delayed diagnosis, ineffective therapy, and poor overall outcomes. Furthermore, Black women have a greater incidence of HR-negative (particularly triple-negative) breast tumors, which are more aggressive and resistant to traditional treatments. This biological difference, together with inequities in access to care, contributes to the gap in death rates.

In addition, social determinants of health such as institutional racism, socioeconomic inequities, and implicit bias in healthcare might exacerbate racial disparities in breast cancer. Such factors are impactful to excellent care, treatment options, and overall health outcomes. To address these gaps, a multimodal approach is required including improving access to healthcare, doing targeted research on triple-negative breast cancer, and tackling systemic inequities to establish an equitable healthcare system [9].

This study focuses on large racial and ethnic differences in breast cancer outcomes in the United States. Despite advances in diagnosis and therapy, black and hispanic women have inferior outcomes than white women. Such inequality is the main cause of a complex combination of causes like systematic racism, socioeconomic inequities, and a lack of access to competent healthcare. Furthermore, biological differences have a higher incidence of aggressive breast cancer subtypes among Black and Hispanic women, leading to a worse prognosis. Furthermore, the underrepresentation of racial and ethnic minorities in clinical trials reduces our understanding of how medicines affect different groups. To address these disparities, a multifaceted approach is required, including improving access to healthcare, conducting targeted research on the specific needs of different racial and ethnic groups, increasing clinical trial diversity, and addressing systemic inequalities to create a more equitable healthcare system [10].

**Dataset used:**
This study analyzes dataset [1] from the Surveillance, Epidemiology, and End Results (SEER) Program to look into differences in breast cancer survival. NCI (National Cancer Institute) manages the SEER program, a comprehensive cancer surveillance system and is largely regarded as vital for cancer epidemiology research. It gathers incidence of cancer data from cancer registries based on population serving roughly a percentage of 47.9 in the US population. SEER registries gather data regarding patients' demographics, the initial tumor premises, tumor morphology, stage at diagnosis, and early stages of treatment and they follow up with patients for vital status.

The SEER dataset contains 96,789 entries and 12 features, which include factors like sex, year of diagnosis, age, race, median household income, rural-urban continuum codes, cancer stage, and cause of death. Significant demographic details such as "Race record (W, B, AI, API)" and "Race and origin record (NHW, NHB, NHAIAN, NHAPI, Hispanic)" allow for more extensive analysis of disparities in race. Socioeconomic facts like "Median household income inflation adjusted to 2022" and geographical identifiers like "Rural-Urban Continuum Code" allow for an analysis of survival disparities affected by geography and economic status.

The dataset's precise and complete records allowed for an analysis of survival differences by race and geographic region, bringing facts about how social factors of health affect breast cancer outcomes. To guarantee compliance with ethical principles, only anonymous patient records were analyzed, and all analyses were carried out in accordance with the SEER data use agreement.

The SEER dataset is publicly available and was obtained from the SEER program's official website (https://seer.cancer.gov/data/). Access to the dataset was allowed following the ratification of the data-use agreement, ensuring that all research activities met the highest standards of data integrity and privacy. Its comprehensive coverage of population demographics enabled the incorporation of both racial and geographic views into survival analyses. This dataset offered a necessary foundation for examining the influence of systemic inequities on breast cancer outcomes, which is a major emphasis of this research.

### III. Methodology

#### A. Dataset Preprocessing

This study's dataset came from the Surveillance, Epidemiology, and End Results (SEER) Program, which collects extensive cancer statistics based on population registries. The collection of data contains 96,789 entries and 12 variables, including sex, year of diagnosis, age, race, income, location, cancer stage, and survival details. Demographic indicators such as "Race record (W, B, AI, API)" and "Race and origin record (NHW, NHB, NHAIAN, NHAPI, Hispanic)" had been utilized for analyzing survival disparities among racial groups, while socioeconomic variables such as median household income and rural-urban continuum codes backed geographic and economic analyses.

Preprocessing involved several steps to ensure data quality:

- **Data Cleaning:**
    - Data with missing or incomplete survival times, unclear staging, or an undetermined race/ethnicity were removed.
- **Feature Engineering:**
    - For standardized analysis, continuous data such as survival time) were converted to years rather than months.
    - Binary event variables were established to indicate whether the death was due to breast cancer or another reason.
- **Standardization and Encoding:**
    - Categorical data, such as race and tumor stage, has been encoded for computational modeling.
    - Exceptions and abnormalities have been found and addressed to ensure data integrity.
- **Dataset Filtering:**
    - Incomplete records in crucial categories such as "survival years" and "SEER cause-specific death classification" were eliminated.

#### B. Survival Analysis Techniques

- **Exploratory Data Analysis (EDA):** EDA was applied to analyze vital variables, identify patterns, and analyze patterns in survival data. Histograms and bar charts gave an overview of the distribution of demographic and clinical variables.
- **Kaplan–Meier Survival Analysis:** The survival probabilities were computed and shown using Kaplan-Meier survival curves. Comparisons between groups were made possible through categorization based on significant demographic criteria such as race and geographic region.
- **Log-rank Test:** Used for analyzing statistical variations in survival distributions across categories discovered through Kaplan-Meier analysis.
- **Cox Proportional Hazards Model:**
    - A Cox Proportional Hazards model was employed to determine the effect of demographic and clinical factors (such as race, rural-urban codes, and tumor stage) on survival.
    - Hazard ratios (HRs) were computed to determine the corresponding probability of mortality for each predictor.

#### C. Tools and Frameworks

- **Python**:

- Libraries such as pandas, matplotlib, and seaborn were employed for data preprocessing, visualization, and exploratory analysis.
- The lifelines library was used for Kaplan-Meier survival analysis and log-rank testing.

**D. Ethical Accountability**

The SEER dataset includes anonymous patient records, which ensures ethical compliance. Every analysis followed the SEER data-use agreement, which guaranteed privacy and data integrity throughout the research process

## IV. Results

Using a dataset from a comprehensive cancer registry, the analysis integrates demographic, socioeconomic, and clinical variables to uncover patterns and trends that affect survival.

**A. Demographic Analysis**

The patient demographic breakdown reveals a nearly equal distribution of cancer incidence between genders, with females slightly outnumbering males. It is expected in breast cancer studies because of the highest prevalence of the disease among women. However, distribution among racial groups showed considerable disparities. Whites constituted the majority with 78.6%, followed by Blacks at 13.2%, Asian/Pacific Islanders at 3.9%, and American Indians/Alaska Natives at 0.3%.

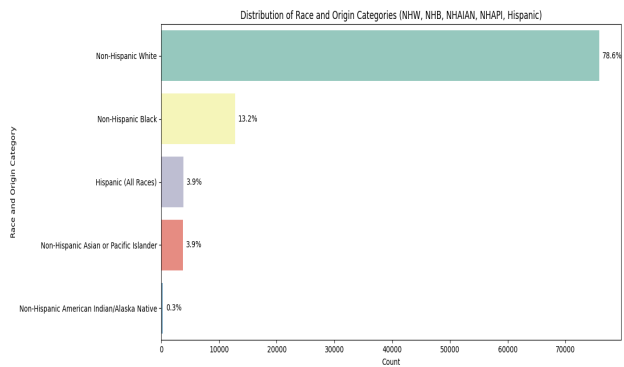

The temporal trends in the dataset from 1975 to 2017 indicated fluctuating diagnosis rates, with notable peaks suggesting possible changes in diagnostic practices, disease awareness, or data collection methodologies. These fluctuations could also be influenced by public health initiatives aimed at improving cancer screening and early detection.

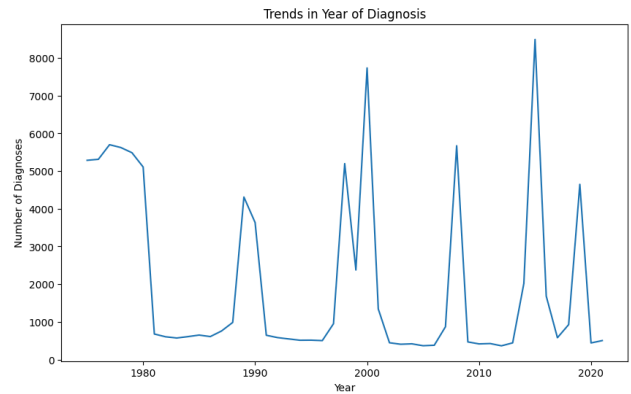

**B. Socioeconomic Factors**

Analysis of the median household income, adjusted for inflation to 2022 values, highlighted significant variances in economic backgrounds among the patients, with a substantial portion of the dataset lacking complete income data. This poses a challenge in fully understanding the socioeconomic factors at play. The Rural-Urban Continuum Codes further provided insight into geographic disparities, where patients from more rural areas generally displayed poorer survival outcomes compared to their urban counterparts. This discrepancy underscores the influence of healthcare accessibility and infrastructure, which are often less developed in rural settings.

```
Median household income inflation adj to 2022
Unknown/missing/no match/Not 1990-2022    42121
$75,000 - $79,999                          7519
$70,000 - $74,999                          6883
$85,000 - $89,999                          6716
$80,000 - $84,999                          6332
$65,000 - $69,999                          5576
$90,000 - $94,999                          5559
$100,000 - $109,999                        3306
$110,000 - $119,999                        3207
$60,000 - $64,999                          2714
$95,000 - $99,999                          2425
$55,000 - $59,999                          1653
$50,000 - $54,999                          1320
$120,000+                                  1054
$45,000 - $49,999                            68
< $40,000                                     1
```

### C. Clinical Stages and Survival Analysis

The SEER combined summary stages and historic stages offered a detailed view of cancer progression at diagnosis, which is crucial for survival analysis. Patients diagnosed at localized stages exhibited significantly higher survival rates than those diagnosed at regional or distant stages, emphasizing the critical role of early detection. The Kaplan-Meier survival curves, stratified by race and sex, showed that non-Hispanic White females consistently had better survival probabilities than other racial groups, with Hispanic and Black patients experiencing the lowest survival rates.

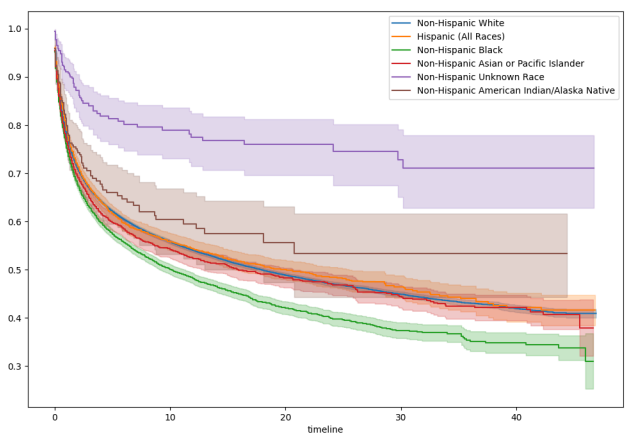

The survival analysis employed Kaplan-Meier estimators and Cox proportional hazards models to evaluate survival times and the impact of covariates on survival probabilities. These models are well-regarded in clinical statistics for their robustness in handling censored data and their capability to adjust for various confounders. However, the assumption of proportional hazards in Cox models may not hold across all stratifications, potentially skewing the results in certain demographic segments.

### D. Implications of Findings

This study reveals significant racial disparities in breast cancer survival rates, demanding focused action to address these inequalities. Notably:

Native Americans demonstrate a concerning 17% lower survival rate compared to White patients, highlighting a critical need for targeted interventions within this population. Black patients show a minor (1%) reduction in risk compared to White patients. While this difference is small, it warrants further investigation to understand the underlying factors.

These findings underscore the urgency for public health initiatives and clinical practices that enhance screening programs, particularly in underserved communities with lower survival rates, ensuring early detection and timely treatment. Increase breast cancer awareness - culturally sensitive education campaigns are crucial to improve knowledge and encourage early detection among all racial groups. Improve access to advanced healthcare - addressing barriers to specialized treatment centers and ensuring equitable access to innovative therapies.

## V. Conclusion, Future Work, and Limitations

### A. Conclusion

The extensive analysis of breast cancer survival disparities reveals a complex interplay of demographic, socioeconomic, and clinical factors. While advancements in healthcare have generally improved cancer survival rates, these benefits are not equally distributed across all population segments. The significant racial and geographic disparities observed underscore the urgent need for tailored healthcare interventions and policies that ensure equitable access to cancer care services.

**B. Future Work**

Future research would focus on integrating more comprehensive datasets that include richer socioeconomic details, lifestyle factors, and genetic markers to better understand the multifaceted influences on breast cancer survival. Longitudinal studies would be particularly valuable in assessing the impact of changing healthcare policies and technological advancements over time. Moreover, employing advanced statistical models such as machine learning could enhance predictive accuracy and uncover more nuanced insights into the factors driving disparities in cancer survival.

**C. Limitations**

This study is not without limitations. The incomplete data on socioeconomic status and the broad categorization of race may obscure deeper insights into intra-racial disparities. The retrospective nature of the data limits the ability to infer causality and may not accurately reflect recent healthcare innovations. Additionally, the reliance on historic stage data, while informative, does not account for the full spectrum of modern diagnostic and treatment modalities.